\pdfoutput=1

\documentclass[11pt]{article}

\usepackage[final]{coling}

\usepackage{times}
\usepackage{latexsym}
\usepackage{amsmath}
\usepackage[T1]{fontenc}

\usepackage[utf8]{inputenc}

\usepackage{microtype}

\usepackage{inconsolata}

\usepackage{graphicx}

\title{MARSAD: A Multi-Functional Tool for Real-Time Social Media Analysis}

\author{%
Md. Rafiul Biswas$^1$, 
Firoj Alam$^2$, 
Wajdi Zaghouani$^3$\\
$^1$Hamad bin Khalifa University, Qatar, 
$^2$Qatar Computing Research Institute, Qatar \\
$^3$Northwestern University in Qatar, Qatar,
}
\begin{document}
\maketitle
\begin{abstract}
MARSAD is a multi-functional natural language processing (NLP) tool designed for real-time social media monitoring and analysis, with a focus on the Arabic-speaking world. It empowers researchers and users without coding expertise to analyze both live and archived social media data, generating detailed visualizations and reports across dimensions such as sentiment analysis, emotion analysis, propaganda detection, fact-checking, and hate speech detection. MARSAD also offers secure data scraping capabilities using API keys for public social media data. MARSAD’s backend architecture combines flexible document storage with structured data management, enabling efficient handling of large and multimodal datasets. The frontend is designed to be user-friendly, allowing for seamless data upload and interaction. This paper outlines the architecture, key features, and future enhancements of MARSAD, positioning it as an essential tool for social media analytics.\footnote{A video demonstrating the system can be found at \url{https://youtu.be/ Bln8iSvwtWg?si=fLSfI4cpmbNSks33.}}

\end{abstract}
\section{Introduction}
MARSAD aims to empower users—ranging from researchers to media analysts—with insights into public opinion and social media dynamics. By providing in-depth analysis of online conversations, the tool helps identify influential figures, track trending topics, and find the sentiment of discussions. This tool leverages advanced Natural Language Processing (NLP) technologies and traditional software development technologies. There are some existing tools like \verb|Communalytic|\footnotemark{} \footnotetext{https://communalytic.org/}that allow users to download data and perform different analyses such as sentiment, toxicity, and topic analyzer and others. They use text embedding from pre-trained transformer-based models to find topics dynamically without human involvement. The core of the process involves converting the text (social media posts, in this case) into embeddings, which are numerical representations of the text's meaning. Commonly, pretrained transformer-based models such as BERT \cite{devlin2018bert}, RoBERTa \cite{liu2019roberta}, Sentence-BERT (SBERT) \cite{reimers2019sentence}, or word2vec are used to generate these embeddings. Then, they classify the text by using clustering and labeling them. However, this application mainly focuses on English, French, and German languages.

European News brief  \verb|EMN News Brief|\footnotemark{}\footnotetext{https://emm.newsbrief.eu} accumulated news sources from online news portals and provides a summary of results that got the highest attraction among news sources. However, they focus only on news channels and do not focus on social media channels. Another tool, social media content \verb|Repear|\footnotemark{}  \footnotetext{https://reaper.social/}, scrapes data from social media. By providing access tokens, API keys, and search parameters, this tool enables downloading public content available on social media. NLP tool  \verb|Social Media Sentiment Visualization|\footnotemark{}\footnotetext{https://www.csc2.ncsu.edu/faculty/healey/social-media-viz/production/} can perform real-time analysis, but it is limited to Reddit and Threads. Also, it does not allow the uploading of user files and analysis of the data. \verb|Tanbih|\footnotemark{} \footnotetext{https://tanbih.qcri.org/}  allows users to identify propagandist content \cite{shah2024mememind}, fact-checking \cite{walter2020fact}, bias detection \cite{rodrigo2024systematic}, and clustering, especially in the Arabic language article.

What sets MARSAD apart is its exclusive focus on supporting the Arabic language, and it integrates the significant types of analysis (e.g., sentiment, propaganda, fact-checking) considerably in a single platform \cite{biswas2024mememind,alam2024propaganda}. MARSAD is capable of handling multiple user requests and performing several tasks simultaneously. Content that does not require an API key, such as Reddit and online news portals, can be analyzed in real-time and provide different kinds of visualizations and summary results. By providing the API key and access token obtained from social media (e.g., Facebook, Twitter, YouTube), MARSAD will work as a data crawling platform, same as \verb|CrowdTangle.|\footnotemark{}\footnotetext{https://transparency.meta.com/researchtools/other-datasets/crowdtangle} Moreover, it includes the integration of five major regional Arabic dialects, making it the first social media monitoring tool to offer such comprehensive linguistic coverage in Arabic. Through its state-of-the-art NLP tools, MARSAD handles the linguistic diversity of the region, accommodating variations in dialects and enabling more accurate and culturally relevant analysis. Additionally, we plan to integrate Arabic-named entity recognition \cite{qu2023survey}, enabling the identification of proper names, geographical locations, and organizations within social media posts \cite{lozano2017tracking}.

\section{MARSAD Architecture}
\begin{figure}
    \centering
    \includegraphics[width=1\linewidth]{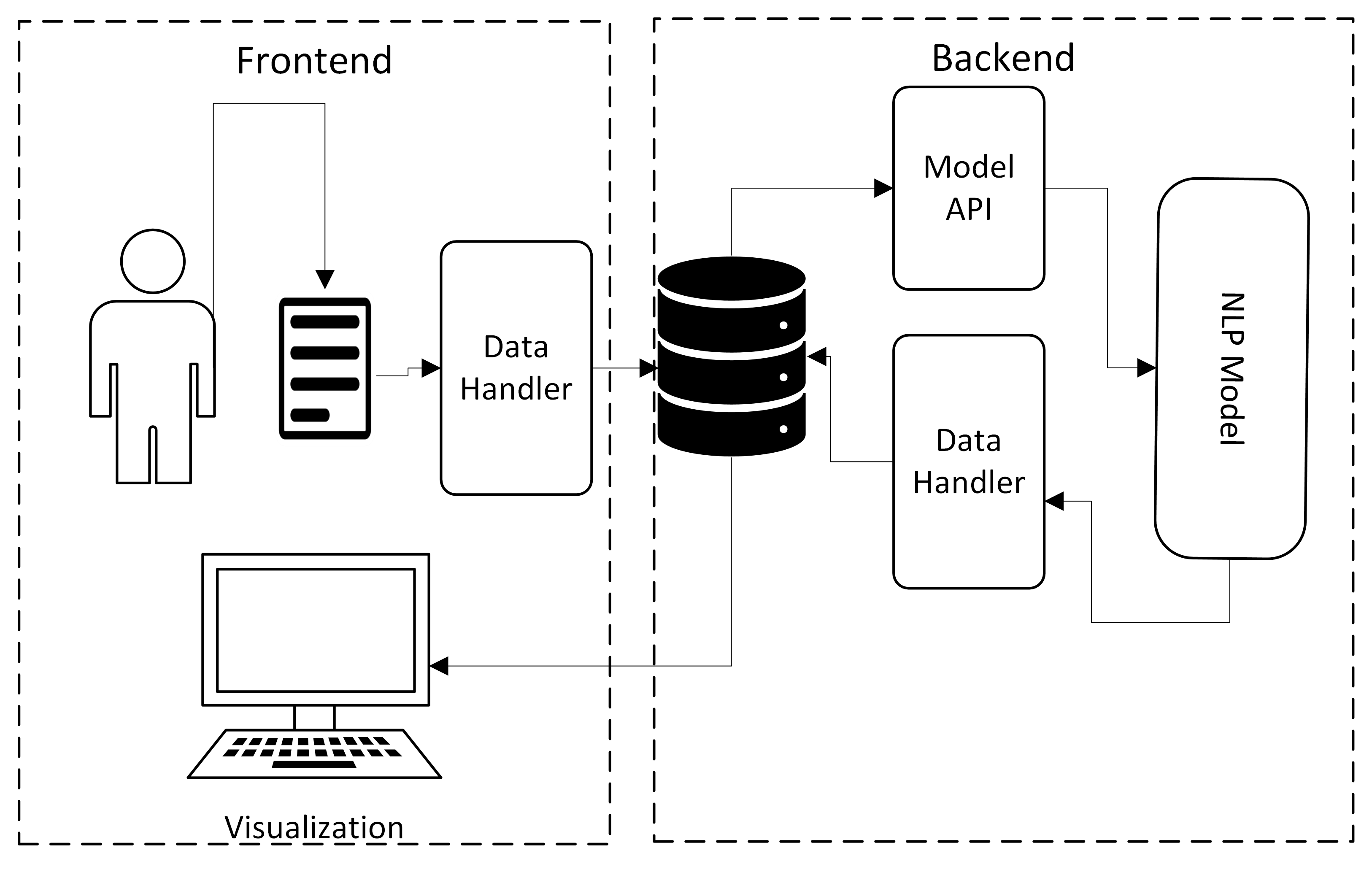}
    \caption{System Architecture}
    \label{fig:architecture}
\end{figure}

MARSAD architectural system can be divided into two major part: i) frontend part ii) backend parts. Figure \ref{fig:architecture} depicts the overview of the system architecture. 

\textbf{Frontend}:
In the frontend, the system interacts directly with the end users, who upload datasets that can include text, images, or multimodal data (combinations of text and images). The system is designed to accept various data formats, such as CSV, TSV, and JSON/JSONL. Once the data is submitted, it is validated with the structure and integrity of the uploaded data. It checks for missing values, correct field types, and schema compliance. It is performed by the Data Handler framework, by the Data Handler framework, which has been developed using modern JavaScript frameworks and libraries, including Next.js for server-side rendering, Flux for unidirectional data flow, and Node.js for server-side operations.

After the validation process, the system generates metadata from the input data and stores it in the appropriate databases. The figure shows this flow, where user-provided data moves through the frontend, is processed by the Data Handler, and then transferred to the backend for further operations.

Once the data is successfully processed and stored, it is made available for visualization. The visualization component allows users to explore and analyze the results through a user-friendly interface, likely including graphical representations of the analyzed data.

\textbf{Backend}: In the backend, the system manages the complex processing, analysis, and storage tasks, ensuring efficient handling of large volumes of data. The data is stored in a \textbf{hybrid database} architecture utilizing both MongoDB \cite{bradshaw2019mongodb}  for flexible, schema-less document storage and PostgreSQL for structured, relational data management. This combination allows the system to scale effectively while handling diverse datasets, including multimodal, structured, and unstructured data.

A core component of the backend is the \textbf{Model API}, which serves as the interface between the application and the underlying NLP models. This API is responsible for coordinating requests for various machine-learning models, including sentiment analysis, propaganda detection, and other NLP tasks. These models are hosted in a containerized environment (e.g., using Docker) to ensure scalability, isolation, and seamless deployment of different versions \cite{docker2020docker}.

The Model API leverages modern deep learning frameworks such as TensorFlow, PyTorch, and Transformers, enabling it to load, fine-tune, and infer from pre-trained models like AraBERT, MARERT \cite{abdul2020arbert},  CamelBERT \cite{inoue-etal-2021-interplay}, or custom models for specialized tasks like propaganda detection. It supports asynchronous processing to handle high-throughput requests and can manage multiple models simultaneously, ensuring low-latency responses and robust performance even under heavy load.

Additionally, the backend side includes \textbf{task queues} \verb|(e.g., Celery)|\footnotemark{}
\footnotetext{https://github.com/celery/celery} to manage long-running or resource-intensive operations like model inference or dataset processing. These queues allow the system to prioritize tasks and manage concurrency efficiently, enabling real-time user interactions with the frontend while delegating complex analyses to the background.

Security and data integrity are maintained through the use of authentication and authorization mechanisms (e.g., OAuth) to ensure that only authenticated users can access and manipulate data or trigger model executions. Data is encrypted both at rest and in transit using AES encryption and SSL/TLS protocols, ensuring compliance with data privacy standards. After the analysis is completed, the results are passed back through the system, where they are processed by another Data Handler component, which further organizes the data and stores the results. Finally, the result is ready for visualization on the frontend. 
\section{Prototype Demonstration}
There are currently four main features in the MARSAD: i) Use cases, ii) Analyze archive data, iii) Search online and analyze, and iv) Data download (see Figure 2). In the dashboard (see Figure 2), users can see the status of their files. Users can start analyzing specific types of analysis (e.g., sentiment) by uploading files. Once the analysis is done, it sends a notification to the users that the analysis has been performed and is ready to visualize. Users can export the file with the analysis report.
Moreover, users can modify the output label that was classified by the tool through a self-annotation process. Users can retrain the model again with user-annotated feedback. This helps the model better learn. Users request multiple analyses at the same time. However, only one analysis is performed at a time, and the other remains in the queue due to the prototype hardware limitation. 
\begin{figure}
    \centering
    \includegraphics[width=1\linewidth]{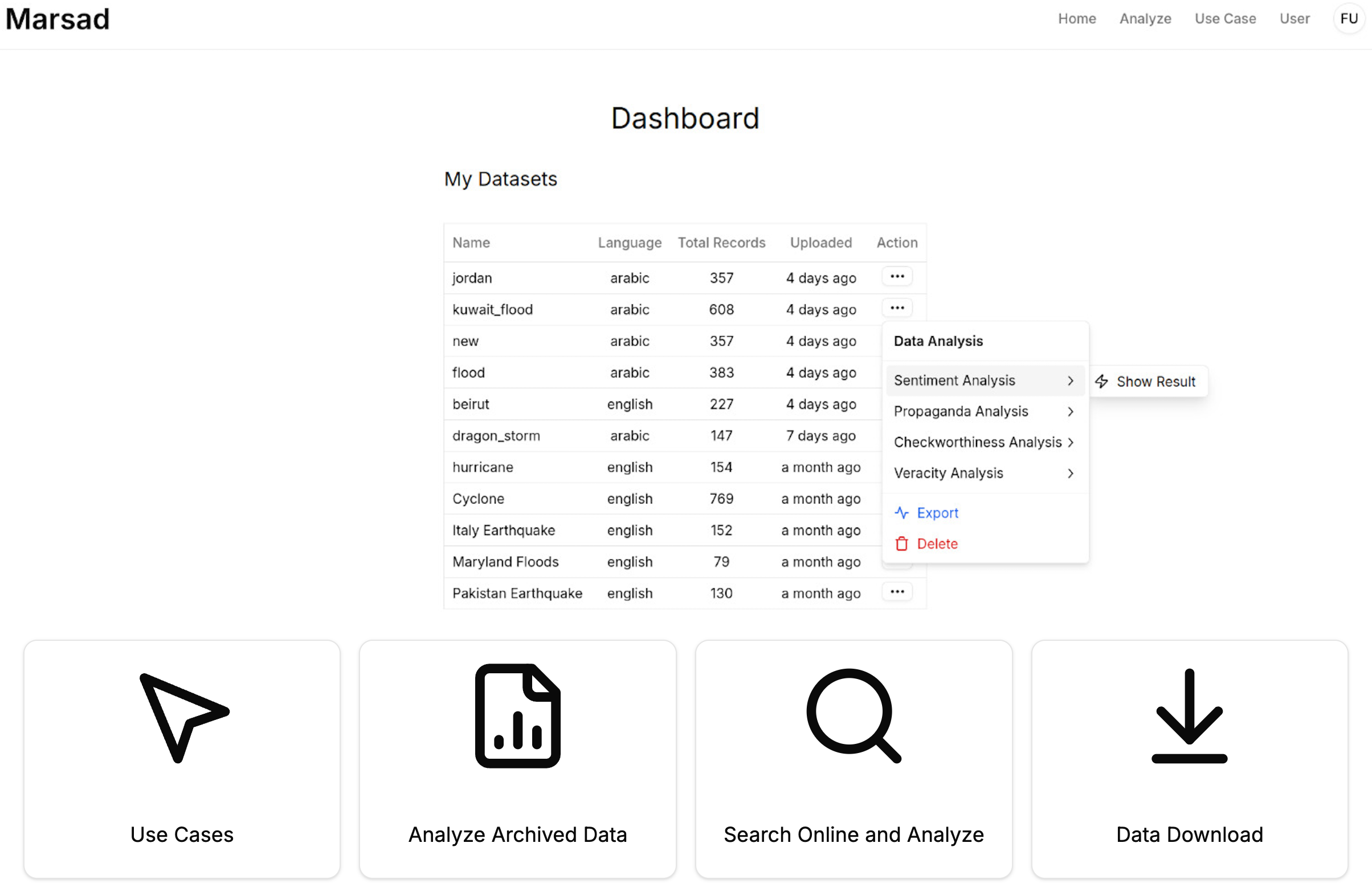}
    \caption{Dashboard}
    \label{fig:dashboard}
\end{figure}

\subsection{Use Cases} MARSAD offers a wide range of use cases (i.e., examples for analyzed results), enabling users to benefit from detailed analysis reports. To support this, we are compiling a large dataset annotated by human experts. These analyzed reports into multiple dimensions are valuable resources for Arabic NLP research and will be made publicly available to the research community under a {Creative Commons license}. 

For this purpose, we have collected only publicly available datasets and anonymized them to prevent the disclosure of any private entities. Next, we annotate data from people in different Arabic dialects. For the annotation tasks, we have employed several annotators through crowd-sourcing. The annotators participated in an intensive two- to three-week training program, which included practical exercises, studying annotation guidelines, and regular meetings to ensure consistency in their work. Once the data is fully annotated, it is processed through our system, and the aggregated results are published for public access. We continuously collect trending data from social media, analyze it through the MARSAD analyzer, and release the findings to the public. These results are freely available for anyone to access and utilize according to their needs.

\subsection{Analyze Archive Data}
Users can upload their datasets—whether collected independently or through external sources—directly to the MARSAD tool. MARSAD supports the upload of both text and image data, allowing for a wide range of analysis tasks. The tool is designed to accommodate various data formats, such as CSV, JSON, or JSONL for text, and standard image formats like JPEG and PNG.

Primarily, MARSAD focuses on processing and analyzing data in Arabic and English, catering to a diverse user base that works with both languages. For text data, MARSAD can handle tasks such as sentiment analysis, propaganda detection, and topic modeling, while image data can be processed for visual sentiment analysis and propaganda detection. For multimodal data—where text and images are combined—the tool will apply specialized models that consider both forms of input to generate more comprehensive insights.

Additionally, MARSAD offers users flexibility in terms of data preprocessing, allowing them to configure settings for tokenization, normalization, and cleaning, especially for text written in Arabic dialects or informal languages. The tool is optimized to handle large datasets and offers real-time feedback on the data upload and analysis processes. Once the data is processed, users can view the results through interactive visualizations and download detailed reports for further study. 

\subsection{Search Online and Analyze}
Users are able to search for online content and analyze data in real-time using MARSAD. This feature allows users to monitor current trends and public sentiment. However, this functionality is available for publicly accessible datasets. There are two ways to perform this task:

Free Access: In this mode, no API key or access token is required. Users select any media platform and provide search terms. MARSAD will then retrieve data from publicly available sources, preprocess it, and perform the analysis. Once completed, the results will be available for visualization.

API-Based Access: For data sources that require an access token, users can provide an API key for specific social media to MARSAD to search the specified data sources. MARSAD will perform the analysis and present the results accordingly.   

\subsection{Data Download}
Users can use the MARSAD tool to download data from social media platforms such as Facebook, X (formerly Twitter), Instagram, YouTube, and more. All they need to do is provide the API key for the respective platform. A tutorial is available to guide users on how to obtain API keys from these platforms and download the data. With no coding required, users can effortlessly gather data from various social media sources.  

\section{Features of MARSAD}
MARSAD is currently conducting several analyses, and more features will be added in the future. For this purpose, we continuously evaluate the performance of the existing language model and transformer-based model and integrate the best model in terms of cost-effectiveness
and performance. 
\textbf{Subtopic and Word Cloud}: MARSAD dynamically identifies subtopics from the user's dataset without requiring prior knowledge of the data. First, we apply the Term Frequency-Inverse Document Frequency (TF-IDF) technique to assess the significance of words in relation to the documents. TF-IDF is calculated as follows:

\[
\text{TF-IDF}(t, d, D) = \text{TF}(t, d) \times \text{IDF}(t, D)
\]

Next, the resulting text embeddings (TF-IDF matrix) are clustered using the K-Means algorithm, where the number of clusters is determined based on the size of the dataset. After clustering, we applied Non-negative Matrix Factorization (NMF) to extract essential words for each cluster. These key terms are then aggregated to represent meaningful subtopics, reflecting the major themes of interest within the data. Additionally, MARSAD generates a word cloud to visually display the most frequently used words in the users' posts, offering an intuitive view of the prominent terms in the dataset.

\textbf{Sentiment Analysis}: We have employed a separate model for language specific to get better results. As there are two types of languages (i.e., Arabic and English), we applied a separate model for sentiment analysis. We use CAMelBERT \cite{inoue-etal-2021-interplay} for Arabic sentiment and google-bert/bert-base-cased \cite{lee2018pre} for English sentiment analysis. Each data will be identified into positive, negative, and neutral sentiments with a score. 

\textbf{Propagandist Content Identification}: We applied \cite{shah2024mememind} techniques to identify whether a text belongs to propagandist content. The techniques can identify propagandist content from text, image, and multimodal data. Moreover, we identified each text's propagandist span and persuasion technique. We continuously add more features like offensive, fact-checking, veracity, and check-worthiness.

\textbf{Sapital analysis}: MARSAD includes functionality to identify and analyze user locations based on the provided data, helping to visualize geographic distribution and trends. By extracting location metadata from user-generated content, such as geotags or inferred locations from text, MARSAD can map where users are located. This spatial analysis allows for a deeper understanding of user demographics, regional preferences, and location-based trends, providing valuable insights into how opinions or sentiments vary across different areas. Figure 3. shows an example of spatial analysis.

\begin{figure}
    \centering
    \includegraphics[width=1\linewidth]{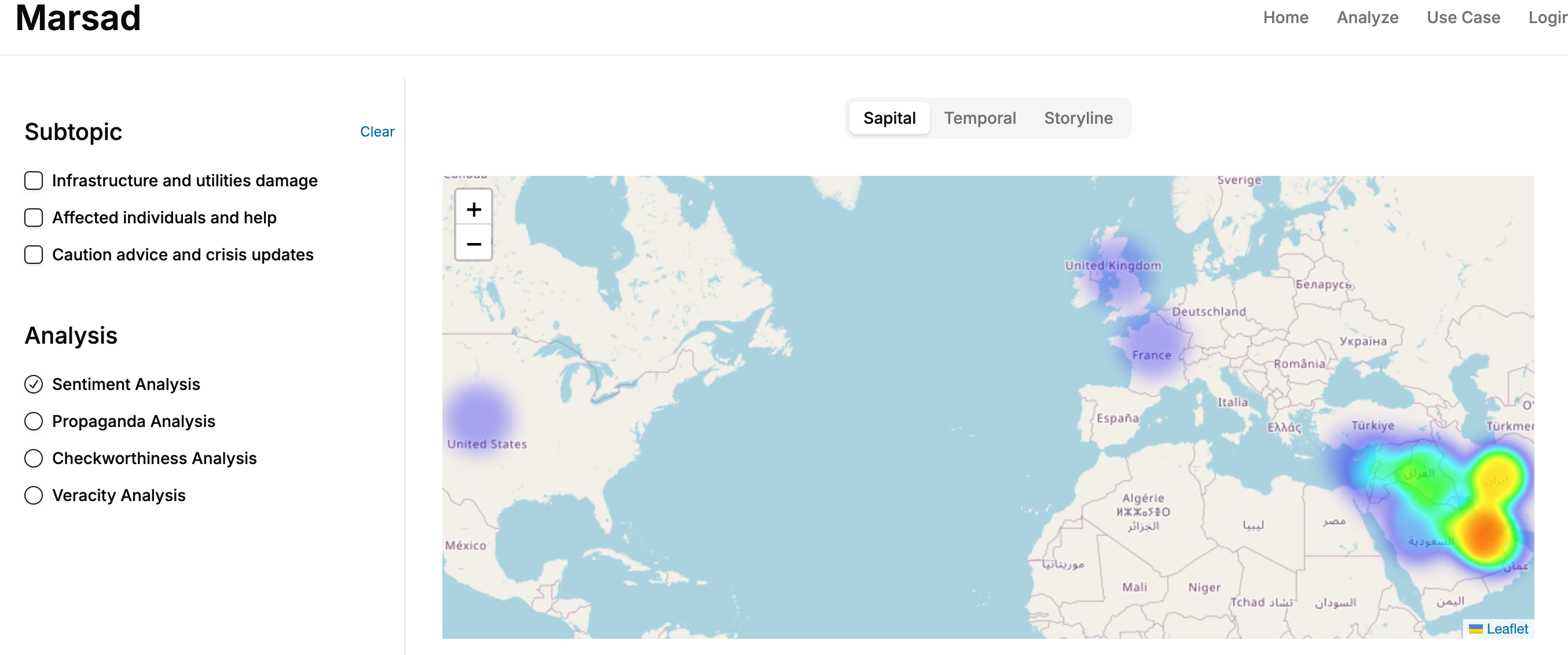}
    \caption{User locations}
    \label{fig:map}
\end{figure}

\textbf{Trending Topics}: Figure 4 shows the trending posts over time. It helps to understand when it was posted and found more public attraction.  This temporal analysis is valuable for identifying patterns in user engagement, such as spikes during critical events, trends around specific dates, or periods of high interaction related to ongoing discussions.

\textbf{Post analysis}: Figure 5 shows that MARSAD enables users to identify label, degree, and location mentions of each post. For instance, a post can be ranked by its sentiment strength, the level of propaganda, or the severity of hate speech. This helps users prioritize content based on its impact or importance, offering deeper insight into each post beyond simple categorization.

\textbf{Network analyzer}: The Network Analyzer feature in MARSAD identifies and visualizes the connections between users who post content and those who interact with it (such as by liking, sharing, or commenting). By mapping out these relationships, the system provides insights into how information spreads within a network and how users influence one another.

\begin{figure}
    \centering
    \includegraphics[width=1\linewidth]{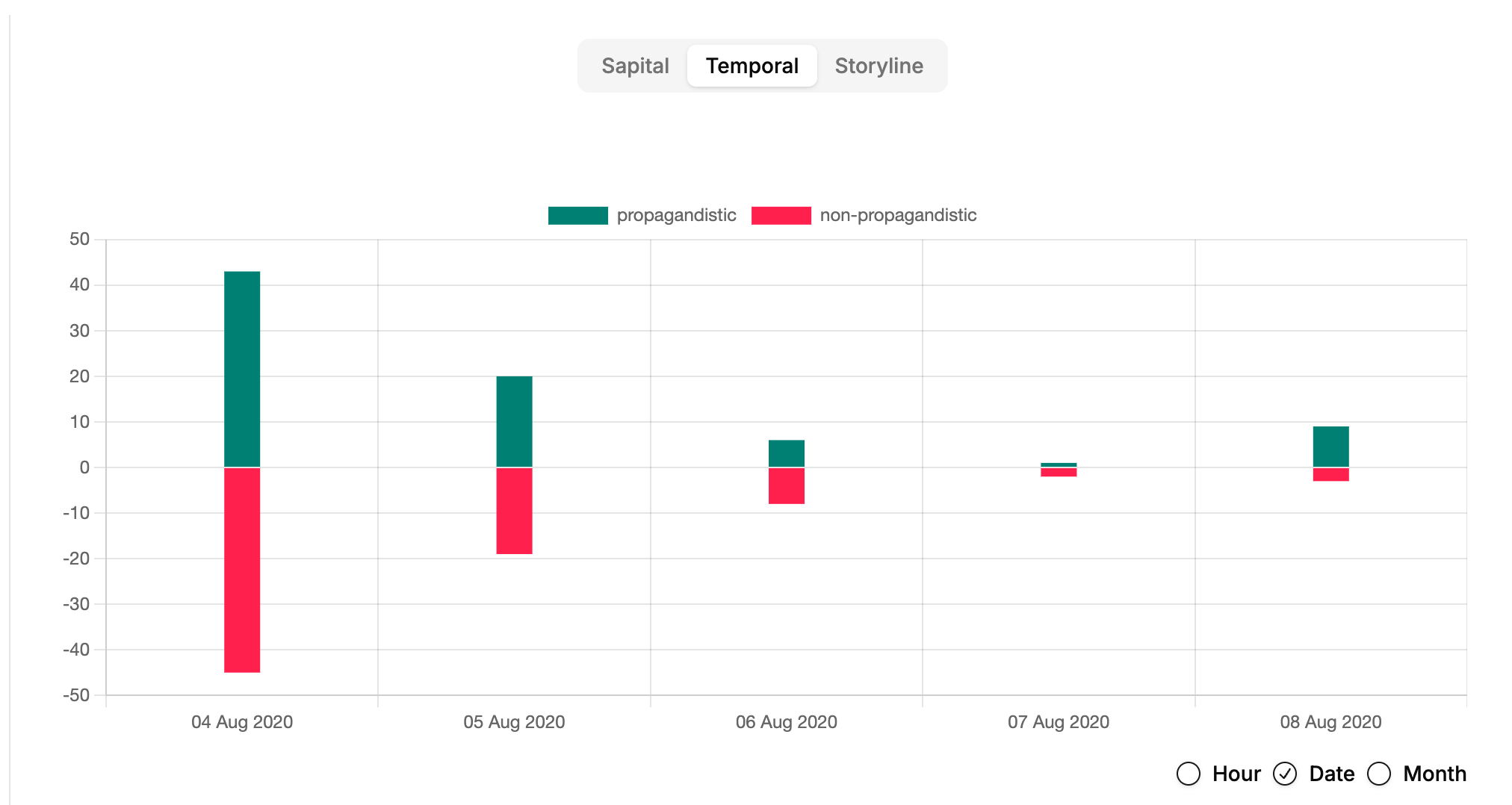}
    \caption{Daily Trends}
    \label{fig:sentiment}
\end{figure}

\begin{figure}
    \centering
    \includegraphics[width=1\linewidth]{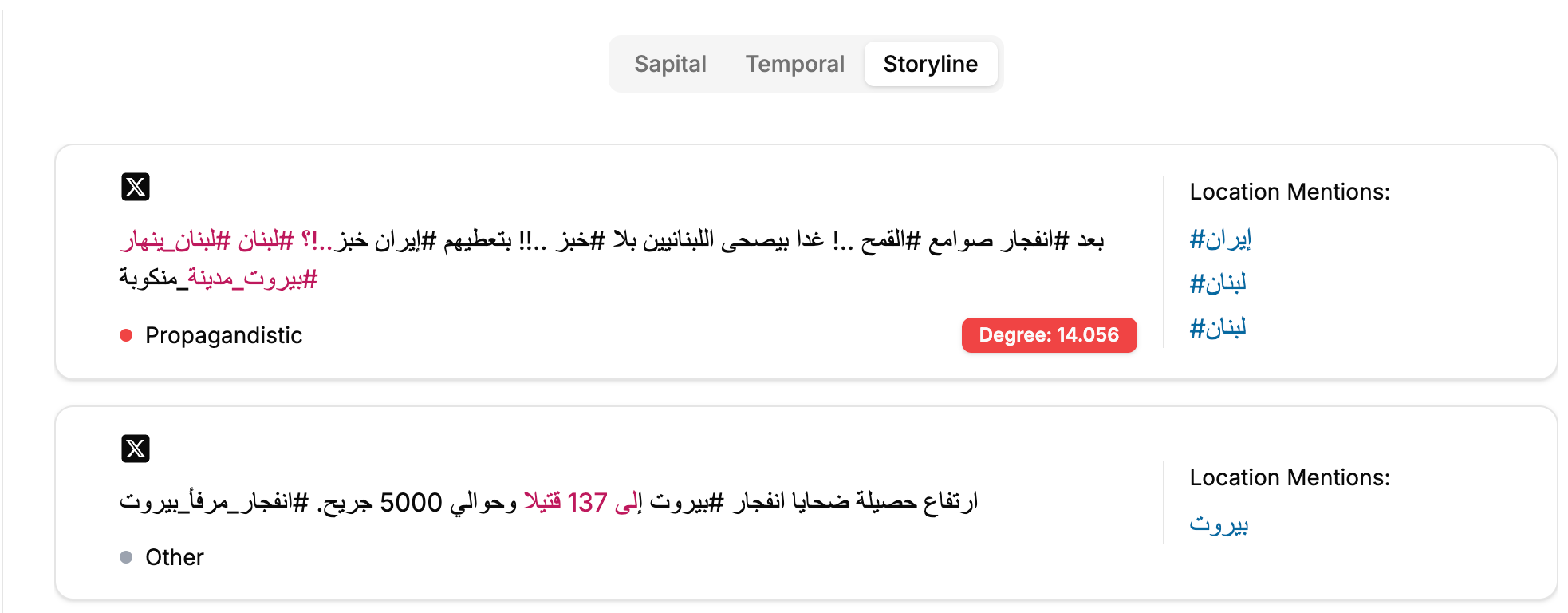}
    \caption{Post analysis}
    \label{fig:post}
\end{figure}

A key contribution of MARSAD is the creation of a large-scale annotated corpus, which will be shared across various sub-projects. This will foster collaboration between different initiatives and provide valuable insights into the broader impact of social media on society. By providing an in-depth analysis of online conversations, the software tool helps identify influential figures, track trending topics, and understand the sentiment of discussions in real-time \cite{el2017web}. In addition to its monitoring capabilities, MARSAD offers detailed analytics that support decision-making and strategy development for businesses, governments, and academic institutions.

\section{Conclusion and Future Works}
MARSAD is a tool that integrates advanced NLP techniques and provides analysis in different dimensions. MARSAD will help the research community, especially non-programmers, analyze social media trends with zero coding knowledge. MARSAD is being developed with continuous feedback to maximize user needs. Currently, most of the NLP models in MARSAD have been developed using the transformer model. In the future, large language models will be integrated to enhance user experiences. Thus, it could deliver a valuable resource for Arabic social media content for those seeking to understand and engage with the region's online discourse. 

\section{Ethical Considerations}
In developing and releasing MARSAD, we are mindful of multiple ethical considerations. The tool is designed for research and educational purposes, focusing on the analysis of online and social media trends, especially the detection of propaganda and hate speech. It is crucial that MARSAD is used responsibly and in line with applicable laws, data privacy regulations, and ethical standards.

Users must ensure that the tool is employed for constructive purposes—such as identifying and mitigating harmful content—and not for amplifying or spreading such content. We strongly encourage users to exercise caution and responsibility in interpreting and sharing the results generated by the tool. Additionally, MARSAD employs machine learning models, including large language models (LLMs), which may introduce inherent biases from the data on which they were trained. To mitigate these risks, we have implemented continuous evaluation processes and encourage users to report any biases or inaccuracies found. Ensuring transparency and accountability is a key objective of this tool.

\section{License}
The MARSAD tool will be released under the \textit{Creative Commons Attribution 4.0 International License (CC BY 4.0)}. This license allows users to share (copy and redistribute) and adapt (remix, transform, and build upon) the tool in any medium or format, provided appropriate credit is given to the original creators. Users must link to the license, indicate if changes were made, and may not apply legal terms or technological measures that legally restrict others from doing anything the license permits. Full details of the license can be found at \url{https://creativecommons.org/licenses/by/4.0/}.

\section{Acknowledgement}
The MARSAD tool is funded by grant NPRP 14C-0916-210015 from the Qatar National Research Fund, part of Qatar Research Development and Innovation Council (QRDI).

\section{Appendix}
The current version of the MARSAD can be found in the following URL \url{https://youtu.be/Bln8iSvwtWg?si=fLSfI4cpmbNSks33}.

\bibliography{custom}
\end{document}